\documentclass[sn-mathphys-ay]{sn-jnl}

\usepackage{graphicx}%
\usepackage{multirow}%
\usepackage{amsmath,amssymb,amsfonts}%
\usepackage{amsthm}%
\usepackage{mathrsfs}%
\usepackage[title]{appendix}%
\usepackage{xcolor}%
\usepackage{textcomp}%
\usepackage{manyfoot}%
\usepackage{booktabs}%
\usepackage{algorithm}%
\usepackage{algorithmicx}%
\usepackage{algpseudocode}%
\usepackage{listings}%

\usepackage{amsmath}

\usepackage{cleveref}
\crefname{section}{Section}{Sections}
\crefname{table}{Table}{Tables}
\crefname{figure}{Figure}{Figures}
\crefname{eq}{Equation}{Equations}

\usepackage{colortbl}
\usepackage{caption}

\definecolor{myblue}{rgb}{0.121, 0.466, 0.705} 
\definecolor{mycolor_base}{named}{myblue}
\newcommand{\colorintensity}{20}  
\newcommand{\mycolor}{mycolor_base!\colorintensity}

\usepackage{makecell}
\usepackage{pifont} 
\newcommand{\cmark}{\ding{51}} 
\newcommand{\xmark}{\ding{55}} 
\newcommand{\Tstrut}{\rule{0pt}{2.6ex}} 

\usepackage{graphicx}
\DeclareGraphicsExtensions{.pdf,.png,.jpg,.jpeg}

\theoremstyle{thmstylethree}%

\begin{document}

\title[Article Title]{StableMamba: Distillation-free Scaling of Large State-Space Models for Images and Videos}

\author*[1,2]{\fnm{Hamid} \sur{Suleman}}\email{hsuleman@iai.uni-bonn.de}
\equalcont{These authors contributed equally to this work.}

\author[1,2]{\fnm{Syed Talal} \sur{Wasim}}\email{swasim@uni-bonn.de}
\equalcont{These authors contributed equally to this work.}

\author[3]{\fnm{Muzammal} \sur{Naseer}}\email{muhammadmuzammal.naseer@ku.ac.ae}

\author[1,2]{\fnm{Juergen} \sur{Gall}}\email{gall@iai.uni-bonn.de}

\affil*[1]{\orgname{Universtiy of Bonn}, 
 \country{Germany}}

\affil[2]{\orgname{Lamarr Institute for Machine Learning and Artificial Intelligence}, 

\country{Germany}}

\affil[3]{\orgdiv{Department of Computer Science}, \orgname{Khalifa University}, \orgaddress{\city{Abu Dhabi},\country{United Arab Emirates}}}


\abstract{State-space models (SSMs), exemplified by S4, have introduced a novel context modeling method by integrating state-space techniques into deep learning. Despite their effectiveness, SSMs struggle with global context modeling due to data-independent matrices. The Mamba model addresses this with data-dependent variants enabled by S6 selective-scan algorithm, enhancing context modeling, especially for long sequences. However, Mamba-based architectures face significant parameter scalability challenges, limiting their utility in vision applications.
This paper tackles the scalability issue of large SSMs for image classification and action recognition without relying on additional techniques like knowledge distillation. We analyze the distinct characteristics of Mamba-based and Attention-based models, proposing a Mamba-Attention interleaved architecture that enhances scalability, robustness, and performance. We demonstrate that the stable and efficient interleaved architecture resolves the scalability issue of Mamba-based architectures and increases robustness to common corruption artifacts. Our thorough evaluation on the ImageNet-1K, Kinetics-400, and Something-Something-v2 benchmarks demonstrates that our approach improves the accuracy of state-of-the-art Mamba-based architectures by up to $+1.7$\%.}




\maketitle

\section{Introduction}\label{sec:intro}


\begin{figure}[t!]
    \centering
    \includegraphics[width=0.7\columnwidth]{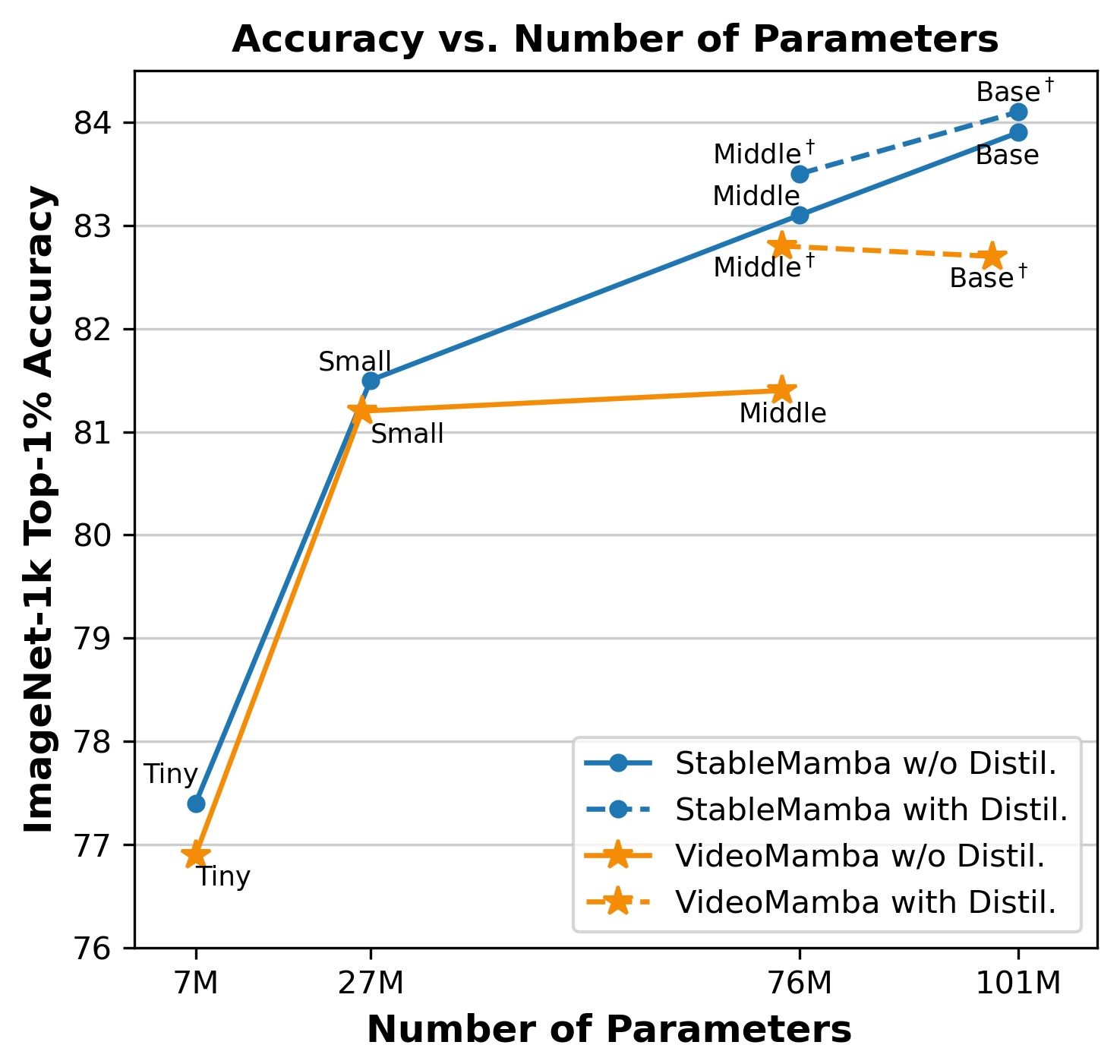}
    \caption{\textbf{Performance comparison with VideoMamba:} We compare the performance of our model with VideoMamba~\citep{li2024videomamba}, both with and without distillation, on IN1K~\citep{deng2009imagenet}.}
    \label{fig:only_intro}
\end{figure}

Various networks have been proposed for both image and video recognition in recent years. These include convolutional neural networks~\citep{alex2012alexnet, he2016resnet, carreira2017I3D, feichtenhofer2019slowfast}, vision Transformers~\citep{dosovitskiy2021vit, arnab2021vivit}, and networks using focal modulation~\citep{yang2022focalnets, wasim2023video-focalnets}. The Attention-based Transformer models have dominated both image and video recognition, either as pure Attention-based models~\citep{liu2021Swin, liu2022swinv2, arnab2021vivit, bertasius2021timesformer, shen2022mtv} or as hybrid models~\citep{li2022uniformer, fan2021multiscale, li2022improved}. 

Recently, State-Space Models (SSMs) such as S4~\citep{gu2022s4} have gained popularity as a new context modeling method. They recurrently model context and bring well-established techniques from state-space modeling to deep large models. However, S4 encountered a problem in terms of modeling global context due to the data-independent nature of the input, state-transition, and output matrices. To mitigate this issue, the Mamba~\citep{gu2023mamba} model introduced the S6 selective-scan algorithm, which uses data-dependent variants of the input and output matrices. This improves the context modeling capabilities, particularly on long sequences, and the approach has been adapted to image tasks~\citep{lianghui2024vim,yue2024vmamba} and in the recent work VideoMamba~\citep{li2024videomamba} to the video domain.   

\begin{figure*}[t!]
    \centering
    \includegraphics[width=1\textwidth]{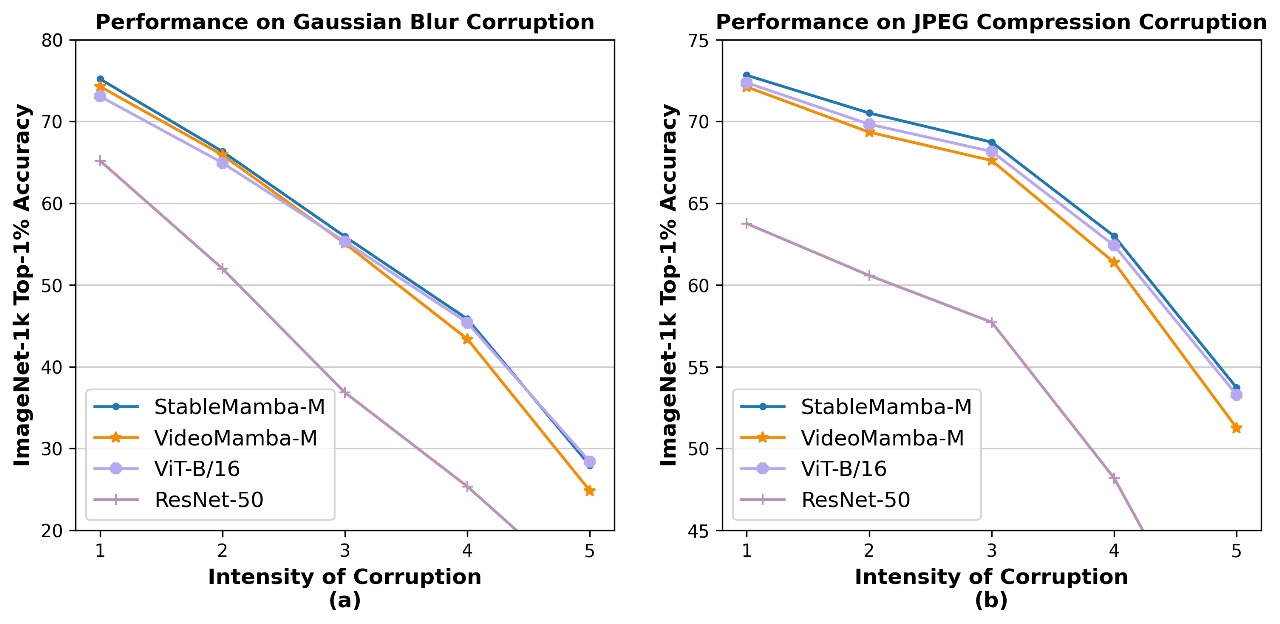}
    \caption{\textbf{(a)} Performance comparison of different networks on Gaussian blur corruption. \textbf{(b)} Performance comparison of different networks on JPEG compression corruption.}
    \label{fig:jpeg_compression}
\end{figure*}

In this work, we investigate the property of vision SSMs, where we focus on VideoMamba~\citep{li2024videomamba} since it is the largest vision SSM architecture and the only that can be applied to videos, and make two key observations. First, VideoMamba does not scale well with the amount of parameters as plotted in Figure~\ref{fig:only_intro}. While the accuracy substantially increases as the number of parameters is increased from 7M (tiny) to 25M (small) parameters, the accuracy only slightly increases if the parameters are increased further to 75M (middle) parameters. To mitigate this issue, \citet{li2024videomamba} proposed to train first a small model and then use the small model as the teacher for training a larger model using distillation. While distillation improves the accuracy of the middle-sized model, it does not solve the underlying problem. Increasing the parameters further to 98M (base) parameters again does not improve the results. 
 
The second observation is the higher sensitiveness of the Mamba-based network to common corruptions and perturbations like image blur or JPEG compression in comparison to vision Transformers as shown in Figure~\ref{fig:jpeg_compression}. Both observations are major limitations for practical applications. We therefore propose a simple yet efficient Mamba-Attention interleaved architecture, termed StableMamba, that resolves both issues. It improves the robustness to common corruptions and perturbations during inference~\citep{hendrycks2019corrImageNet} as shown in Figure~\ref{fig:jpeg_compression} and mitigates the scalability issue without the need of cumbersome workarounds like distillation as shown in Figure~\ref{fig:only_intro}.    
In summary, the main contributions of this paper are:
\begin{itemize}
    \item We analyze the largest Mamba architecture for images and video and present a simple yet efficient Mamba-Attention interleaved architecture.
    \item We show that our approach resolves the scalability issue and increases the robustness to various common corruptions~\citep{hendrycks2019corrImageNet}.
    \item We report improved performance for comparable methods for image classification on ImageNet-1K~\citep{deng2009imagenet} and for action recognition on Kinetics-400~\citep{kay2017k400} and Something-Something-v2~\citep{goyal2017ssv2}.
\end{itemize}

\section{Related Work}\label{sec:related}

\textbf{Image and Video Recognition:} In the last decade, Convolutional Neural Networks (CNNs) have been the primary choice for computer vision tasks. Starting with the introduction of AlexNet~\citep{alex2012alexnet}, the field has seen rapid advancements with notable architectures such as VGG~\citep{simonyan2014vgg}, Inception~\citep{szegedy2015inception}, ResNet~\citep{he2016resnet}, MobileNet~\citep{howard2017mobilenet}, and EfficientNet~\citep{tan2019efficientnet} achieving improved performance on ImageNet~\citep{deng2009imagenet}. Recently, ConvNeXt variants~\citep{liu2022convnext, woo2023convnextv2} and FocalNets~\citep{yang2022focalnets} have updated traditional 2D ConvNets with modern design elements and training techniques, achieving performance comparable to state-of-the-art models. At the same time, the Vision Transformer (ViT)~\citep{dosovitskiy2021vit}, inspired by the Transformer~\citep{vaswani2017attention} for natural language processing, and its variants such as DeiT~\citep{touvron2021deit}, Swin Transformer~\citep{liu2021Swin}, and Swin Transformer V2~\citep{liu2022swinv2} have achieved very good results for image classification.

For Video Recognition, early methods were feature-based~\citep{klaser2008spatio, Laptev2003STinterestPoints, wang2013dense}. Later, the success of 2D CNNs~\citep{alex2012alexnet, simonyan2014vgg, he2016resnet, tan2019efficientnet} on ImageNet~\citep{deng2009imagenet} lead to their application to video recognition~\citep{karpathy2014large, Joe2014beyond, simonyan2014two}. However, these methods lacked temporal modeling capabilities. The release of large-scale datasets such as Kinetics~\citep{kay2017k400} prompted 3D CNN based methods~\citep{carreira2017I3D, feichtenhofer2016spatiotemporal, tran2015C3D}. Since these were computationally expensive, various methods were proposed to mitigate the issue~\citep{feichtenhofer2020x3d, sun2015human, szegedy2016rethinking, tran2018closer, xie2018s3d, li2020tea, lin2019tsm, qiu2019learning, feichtenhofer2019slowfast, duan2020omni, li2020ct-net, wang2021tdn}. When the ViT~\citep{dosovitskiy2021vit} architecture became popular in image recognition, it seamlessly made its way into the video domain. Initial methods used Self-Attention in combination with CNNs~\citep{wang2018nonlocal, wang2020nas, kondratyuk2021movinets} while later works~\citep{liu2021video-swin, arnab2021vivit, bertasius2021timesformer, shen2022mtv, zhang2021vidtr, patrick2021mformer, fan2021multiscale, li2022improved, patrick2021mformer, sharir2021stam} introduced pure Transformer based architectures. More recently, Video-FocalNets~\citep{wasim2023video-focalnets} proposed a Focal Modulation~\citep{yang2022focalnets} extension for videos, while Uniformer~\citep{li2022uniformer} proposed an efficient hybrid architecture for video recognition. Very recently, a key development in this area came with FlashAttention~\citep{dao2022flashattention, dao2023flashattentionv2}, which presents a hardware-aware implementation of the Attention algorithm, mitigating the quadratic compute complexity issue of Attention-based models.

\textbf{State Space Models:} Recently, State-Space Models (SSMs), such as the Structured State-Space Model S4~\citep{gu2022s4}, have been presented as an alternative to Self-Attention~\citep{vaswani2017attention} for efficient modeling of long sequences with linear complexity. Various variants building on the S4 architecture have also been proposed, including S5~\citep{smith2023s5}, H3~\citep{fu2023h3}, and GSS~\citep{mehta2022GSS}. However, the original S4~\citep{gu2022s4} and its variants had a weakness compared to Self-Attention, mainly because they did not have any input dependencies. To mitigate this, \citet{gu2023mamba} proposed the input-dependent state-space model MAMBA alongside an efficient hardware-optimized parallel selective scan mechanism (S6). Various works have been proposed in computer vision applying Mamba to different downstream domains. Two variants were initially proposed for image classification: Vim~\citep{lianghui2024vim} and VMamba~\citep{yue2024vmamba}. Vim proposed an isotropic architecture with a bi-directional scanning variant of Mamba~\citep{gu2023mamba} for effectively scanning the image token sequence. In contrast, VMamba~\citep{yue2024vmamba} proposed a hierarchical architecture with a four-directional scan across all four spatial dimensions. Subsequently, other variants such as LocalVMamba~\citep{huang2024localmamba} had a Swin~\citep{liu2021Swin} style windowed scan while EfficientVMamba~\citep{pei2024efficientvmamba} proposed an atrous-selective scan to improve efficiency. Furthermore, Mamba was also used in various applications in video understanding~\citep{yang2024vivim, li2024videomamba, chen2024video}, image segmentation~\citep{liu2024swin_umamba, ma2024umamba, ruan2024vmunet, gong2024nnmamba}, and various other tasks~\citep{guo2024mambamorph, he2024pan, wang2024graph, guo2024mambair, liang2024pointmamba}. SiMBA~\citep{patro2024simba} uses the Fourier transform with non-linearities to model eigenvalues as negative real numbers in an attempt to improve the training. Similar methods have also been proposed for CNNs~\citep{wang2020high} and Transformers \citep{xiao2021early, touvron2021deit}.
 
A complementary work to ours, VideoMamba~\citep{li2024videomamba}, proposes to use a distillation-based objective to stabilize the training of larger models. However, we show that a simple interleaving of Self-Attention layers within a Mamba-based model is enough to stabilize training for image and action recognition applications and improve robustness against high frequencies in the input.

\section{Limitations of Mamba-based Networks for Visual Recognition}
\label{sec:limitations}

Although Mamba-based networks have shown state-of-the-art performance for image classification~\citep{li2024videomamba, lianghui2024vim} and action recognition~\citep{li2024videomamba}, their training is unstable, which limits the scalability of these architectures. 
For instance, VideoMamba~\citep{li2024videomamba} uses a distillation technique to improve training stability and performance. Since the proposed self-distillation technique requires training smaller model first, it is a cumbersome approach that increases the training cost.

\begin{figure}[t]
    \centering
    \includegraphics[width=0.7\columnwidth]{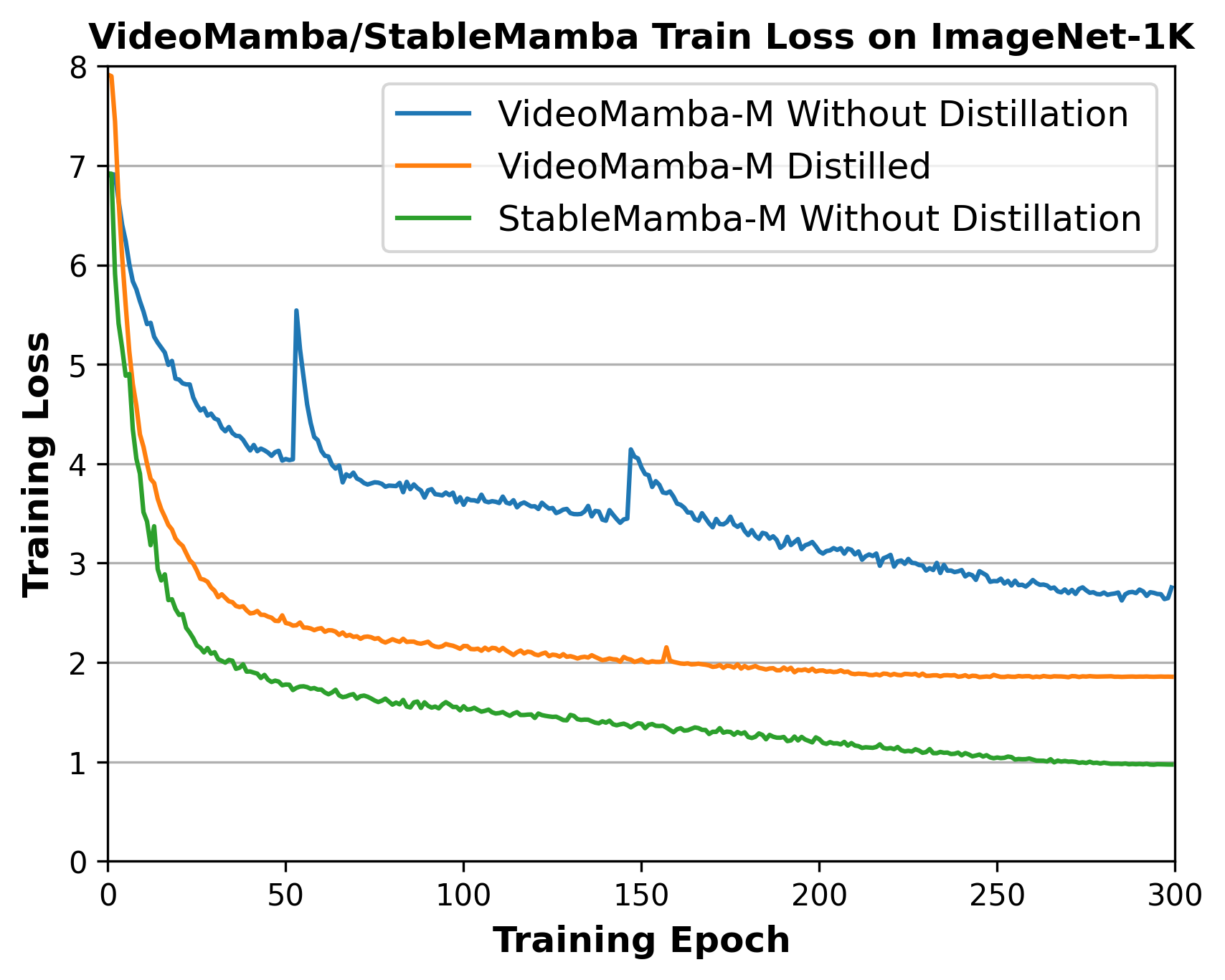}
    \caption{Loss curves obtained from training VideoMamba with and without distillation.}
    \label{fig:videoMamba_training_curves}
\end{figure}

Before we propose our solution to the scalability problem in Section~\ref{sec:method}, we analyze the behavior of pure Mamba-based visual architectures in more detail. We focus on VideoMamba~\citep{li2024videomamba} since it is the largest architecture and the only one that can be applied to video data. VideoMamba trains its tiny and small models with 7M and 25M parameters, respectively, in a conventional setting. However, distillation is used to train it as soon as the parameters are scaled up to the middle model (75M parameters) and base model (98M parameters). The method uses the smaller model as the teacher for the larger middle and base models. This is a departure from the general knowledge distillation where a larger complex model is distilled into a smaller student model~\citep{lieber2024kdistill}. This reversal suggests that the purpose of distillation is not merely to transfer knowledge from a simpler model to a complex one but to stabilize the learning process of the middle and base models. As shown in Figure~\ref{fig:only_intro}, the architecture cannot be scaled beyond 25M parameters without distillation, i.e., the accuracy does not increase further. While distillation improves the accuracy, it does not address the scaling issue since the base model is not better than the middle model. To better understand the impact of distillation on the training, we trained VideoMamba's middle variant with and without distillation. The training curves shown in Figure~\ref{fig:videoMamba_training_curves} indicate the presence of instabilities without distillation. We also present, in Figure~\ref{fig:videoMamba_training_curves}, the loss curve for our StableMamba, which has a stable convergence without distillation. 


Furthermore, in Figure~\ref{fig:jpeg_compression}, we compare the behavior of VideoMamba~\citep{li2024videomamba} with ViT-B\textbackslash16~\citep{dosovitskiy2021vit} under an increasing amount of Gaussian blurring in the input image during inference. For this, we use the images from the ImageNet-C~\citep{hendrycks2019corrImageNet} benchmark, which evaluates the robustness of networks to common corruptions like Gaussian blur. As shown in Figure~\ref{fig:jpeg_compression}(a), VideoMamba~\citep{li2024videomamba} suffers more than the vision Transformer from high intensities of Gaussian blurring. The better robustness of ViT-B\textbackslash16 can be explained by the fact that Transformers tend to focus on lower frequencies in the input image~\citep{naseer2021intriguingProps}.
This observation is further supported by another experiment that examines the behavior of networks under JPEG compression corruption. JPEG compression primarily removes high frequencies as the compression rate increases, although it also introduces tertiary compression-related artifacts as well. The removal of higher frequencies remains the dominant effect. Figure~\ref{fig:jpeg_compression}(b) shows that the VideoMamba is less robust to corruptions of higher frequencies, and addressing this challenge is an important contribution of this paper.

The above-mentioned observations provide enough evidence that it is difficult to scale Mamba models. Using distillation with a smaller model is a workaround to address training instabilities for larger models since it penalizes the larger model for deviating from the smaller one and thus acts as a regularization constraint, but it does not resolve the scalability issue. Furthermore, they are less robust to common image corruptions than vision Transformers. We thus propose an efficient distillation-free solution that mitigates the scalability issue, including training stability issues for large models, and improves the robustness to common image corruptions. Our solution is motivated by the fact that vision Transformers suffer less from these issues and we hypothesize that adding attention blocks to pure Mamba-based visual architectures resolves these issues. We evaluate the effectiveness of this hypothesis in the subsequent sections.

\section{StableMamba for Image Classification and Action Recognition}
\label{sec:method}

Before discussing the StableMamba architecture in Section~\ref{sec:method:arch}, we briefly introduce state-space models in general. 

\subsection{State-Space Models}
\label{sec:method:ssm}

State-space models (SSMs) are inspired by continuous systems in which an input signal $u(t)$ is mapped to a latent state $h(t)$ before being mapped to an output signal $y(t)$. Concretely, a linear ordinary differential equation describes the SSM model:
\begin{equation}
    \begin{split}
    h'(t) &= {\mathbf A}h(t) + {\mathbf B}u(t) \\
    y(t) &= {\mathbf C}h(t) 
    \label{eq:math:ssm}
    \end{split}
\end{equation}
where $h(t)$ is the hidden state, $h'(t)$ is the first derivative, $u(t)$ is the input, and $y(t)$ is the output. $\mathbf{A}$ is the evolution matrix, and $\mathbf{B}$ and $\mathbf{C}$ are the projection matrices of the system.

\textbf{Discretization of State-Space Models:} As mentioned before, Equation~\ref{eq:math:ssm} is valid for continuous time systems. To apply Equation~\ref{eq:math:ssm} on a discretized input sequence $(u_0, u_1, u_2, ...)$ instead of a continuous function $u(t)$, Equation~\ref{eq:math:ssm} must be discretized using a step size $\Delta$ which describes the input time-step resolution.  
 The standard discretization that follows Mamba~\citep{gu2023mamba} is the Zero-Order Hold (ZOH) discretization:
\begin{equation}
\begin{split}
\overline{{\mathbf A}} &= \exp(\Delta {\mathbf{A}}) \\
\overline{{\mathbf B}} &= (\Delta {\mathbf{A}})^{-1} (\exp(\Delta {\mathbf{A}}) - {\mathbf I}) \cdot \Delta {\mathbf{B}} \\
h_t &= \overline{{\mathbf A}} h_{t-1} + \overline{{\mathbf B}} u_t \\
y_t &= {\mathbf C}h_t.
\label{eq:meth:discrete_ssm}
\end{split}
\end{equation}
The difference between S4~\citep{gu2022s4} and Mamba~\citep{gu2023mamba} is the selective scan mechanism that conditions the parameters of ${\mathbf{A}}$,
${\mathbf{B}}$, and ${\mathbf{C}}$ on input. 

\newcommand{\var}{\mathbf}
\newcommand{\optr}{\textsf}
\newcommand{\subs}{\mathrm}

\subsection{StableMamba}
\label{sec:method:arch}

\begin{figure*}[t!]
\centering
    \includegraphics[width=1\textwidth]{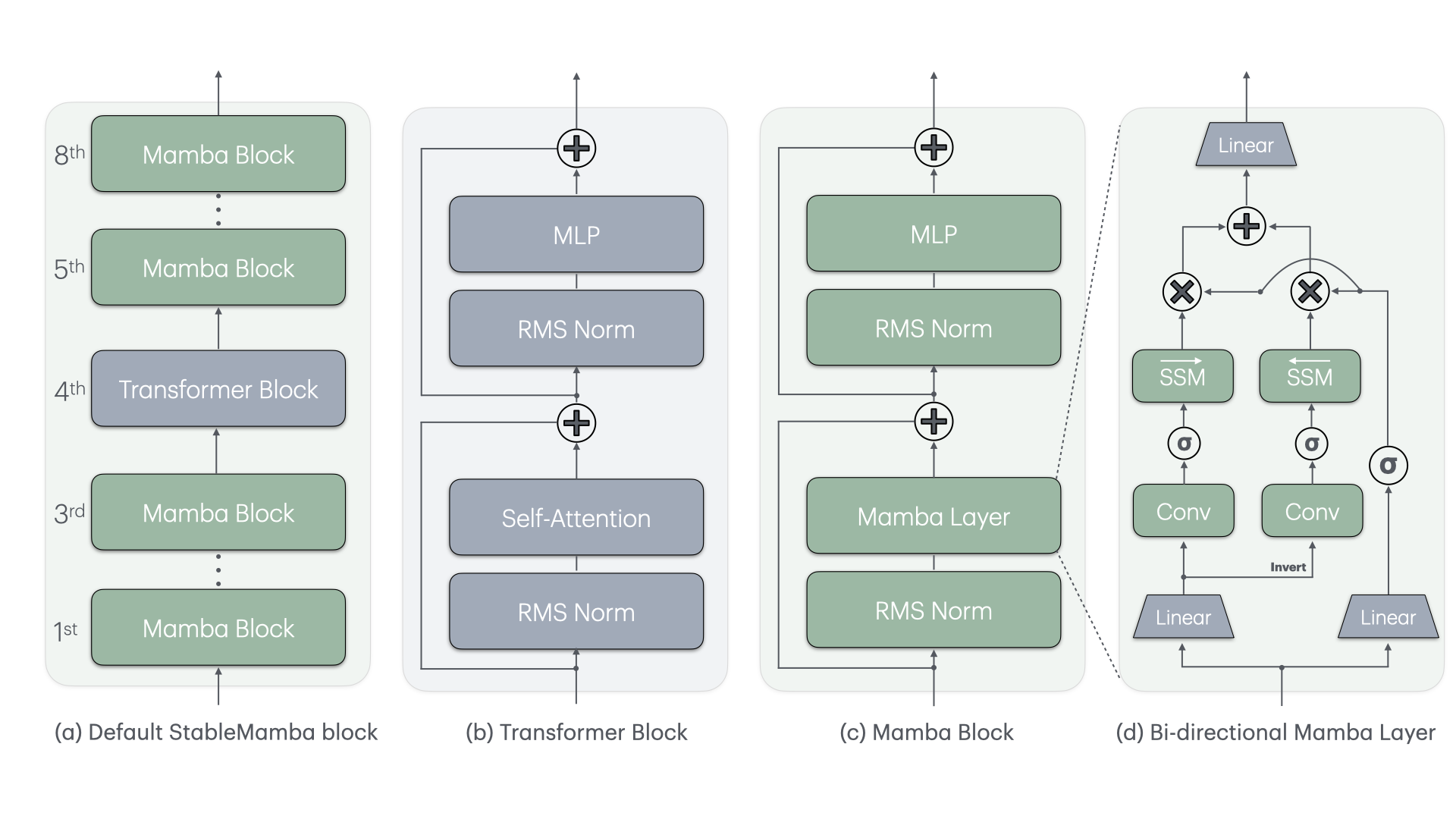}
  \caption{\label{fig:overall_architecture}(a) The overall architecture of the StableMamba model. (b) Anatomy of Transformer block. (c) Anatomy of Mamba block. (d) Anatomy of bidirectional Mamba \textit{layer}.}
\end{figure*}

VideoMamba~\citep{li2024videomamba} uses bi-directional Mamba layers introduced by VisionMamba~\citep{lianghui2024vim} and shown in Figure~\ref{fig:overall_architecture}(d). A bi-directional Mamba block adapts the concept of bi-directional sequence modeling to vision-related tasks. It processes flattened visual token sequences simultaneously using forward and backward state-space models. 

Our architecture consists of stacked StableMamba blocks. Within each StableMamba block are $N$ bi-directional Mamba blocks and $A$ Transformer blocks as shown in Figure~\ref{fig:overall_architecture}(a). The purpose of the Transformer blocks is to stabilize the training and increase the robustness by resetting the focus after several bi-directional Mamba blocks more on lower frequencies. We will evaluate the impact of the number of Transformer blocks in each StableMamba block and the position of the Transformer block within the StableMamba block in Section~\ref{sec:results}. We now describe the two blocks in more detail.   

\textbf{Transformer block:} The Transformer block is detailed in Figure~\ref{fig:overall_architecture}(b). Each Transformer block begins with a Root Mean Square (RMS) normalization layer applied to the input data. It follows a Self-Attention layer where three learnable linear layers $\mathbf{W}^Q$, $\mathbf{W}^K$, and $\mathbf{W}^V$ are used for transforming the input $\mathbf{X}$ into queries ($\mathbf{Q}$), keys ($\mathbf{K}$) and values ($\mathbf{V}$) such that $\mathbf{Q}=\mathbf{X}\mathbf{W}^Q$, $\mathbf{K}=\mathbf{X}\mathbf{W}^K$, and $\mathbf{V}=\mathbf{X}\mathbf{W}^V$. The output $\mathbf{Z}$ of the Self-Attention layer is then calculated as:
\begin{equation}%
\mathbf{Z}= \optr{SOFTMAX}\left (\frac{\mathbf{Q}\mathbf{K}^T}{\sqrt{D_q}}\right )\mathbf{V}
\label{eq:math:atten_qkv}
\end{equation}
where $D_q$ is the dimension of the query. Furthermore, a skip connection is added to the output. Subsequently, another RMS normalization is applied, after which this output is fed to an MLP layer. This constitutes the entire Transformer block shown in Figure~\ref{fig:overall_architecture}(b). The operations can be summarized as:
\begin{equation}
\begin{aligned}
    \var{Z}_{\subs{in}}  &= \optr{PE} + \optr{EMB}(\mathbf{X}) \\
    \var{Z}'_{\subs{out}}  &= \var{Z}_{\subs{in}} + \optr{ATTN}(\optr{RMSNORM}(\var{Z}_{\subs{in}})) \\
    \var{Z}_{\subs{out}}   &= \var{Z}'_{\subs{out}} + \optr{MLP}(\optr{RMSNORM}(\var{Z}'_{\subs{out}}))
\end{aligned}
\label{eq:math:transformer_block}
\end{equation}
where $\mathbf{X}$ is the input to the Transformer block. $\optr{EMB}$ is the convolutional patch embedding and $\optr{PE}$ is the positional encoding as in \citep{dosovitskiy2021vit}. $\optr{RMSNORM}$ is the RMS norm layer and $\optr{ATTN}$ denotes the multi-head Self-Attention layer described in Equation~\ref{eq:math:atten_qkv}. 
The $\optr{MLP}$ is defined by:
%
\begin{align}
    &\optr{MLP}(\optr{RMSNORM}(\mathbf{Z}'_{\subs{out}})) =\\ 
    \nonumber&\quad \optr{GELU}(\optr{RMSNORM}(\mathbf{Z}'_{\subs{out}}) \mathbf{W}_1 + \mathbf{b}_1) \times \mathbf{W}_2 + \mathbf{b}_2.
    \label{eq:ffn}
\end{align}

\textbf{Mamba block:} The Mamba block (Figure~\ref{fig:overall_architecture}(c)) has the same structure as the Transformer block except that it uses a bi-directional Mamba layer instead of a self-attention layer. For brevity's sake, we will call the bi-directional Mamba layer simply as the Mamba layer. The Mamba block performs the following operations: 
\begin{equation}
\begin{aligned}
    \var{Z}'_{\subs{out}} &= \var{Z}_{\subs{in}} + \optr{MAMBA}(\optr{RMSNORM}(\var{Z}_{\subs{in}})) \\
    \var{Z}_{\subs{out}} &= \var{Z}'_{\subs{out}} + \optr{FFN}(\optr{RMSNORM}(\var{Z}'_{\subs{out}})).
\end{aligned}
\label{eq:mamba_mixer}
\end{equation}

Our Mamba block differs from VideoMamba~\citep{li2024videomamba} in that we add an RMS normalization layer and an MLP layer inside the Mamba block. 

The number of parameters of the network can be controlled by the depth of the network and the embedding dimension. We introduce four variations of our model: StableMamba-Tiny has 7M parameters, StableMamba-Small has 27M parameters, StableMamba-Middle has 76M parameters, and StableMamba-Base has 101M parameters.

The complete list of hyperparameters for reproducibility purposes is provided in \cref{tab:hyperparameters}. We use 4 nodes with 4 A100 GPUs (40GB) each for training all of our StableMamba models. 

\begin{table*}[h!]
\centering
\resizebox{1\textwidth}{!}{%
\begin{tabular}{lcccc}
\toprule
 \textbf{StableMamba Training Recipe}
 \\T=Tiny, S=Small, and M=Medium  \\
\midrule
Dataset & IN1K &  K400 & SSv2\\
\midrule
Epochs   & 300 & 70(T), 50(S,M)   &  35(T), 30(S,M)  \\
\midrule
     Batch size & 128 & 32(T)/16(S,M) & 32(T)/16(S,M) \\
     Optimizer & AdamW & AdamW  & AdamW \\
     Optimizer momentum & $\beta_1=0.9,\beta_2=0.999$ & $\beta_1=0.9,\beta_2=0.999$ & $\beta_1=0.9,\beta_2=0.999$\\
     Learning rate       & 5e-4 & 4e-4(T,S), 2e-4(M) & 4e-4 \\
     Minimum learning rate  & 1e-5 & 1e-6 & 1e-6 \\
     Scheduler           & cosine & cosine & cosine \\
     Weight decay        & 0.1(T), 0.05(S,M) & 0.1(T), 0.05(S,M) &  0.1(T), 0.05(S,M) \\
     Warmup epochs       & 5 (T,S), 30(M) & 5 & 5 \\
     Trans. to Mamba blocks & $1:7$ & $1:7$  & $1:7$ \\
\midrule
     Label smoothing & 0.1 & 0.1   &  0.1 \\
     Drop path       & 0(T), 0.15(S), 0.5(M) &  0.1(T), 0.35(S), 0.8(M)   &  0.1(T), 0.35(S), 0.8(M)  \\
     Repeated aug.    & Yes(T),   No(S,M) & 2  &  2 \\
\midrule
     Input size    & $224^2$ & $16\times224^2$  &  $8\times224^2$  \\
     Patch size    & 16 & 16 & 16\\
     Rand. aug.  & (7, 0.25)(T), (9, 0.5)(S,M) & (7, 0.25)(T), (9, 0.5)(S,M)  & (7, 0.25)(T), (9, 0.5)(S,M)  \\
     Mixup prob.   & 0.8 & 0.8  &  0.8 \\
     Cutmix prob.  & 1.0 & 1.0  & 1.0 \\
 \bottomrule 
 \end{tabular}}
\caption{Hyperparameters for StableMamba. Note that StableMamba-B (Base) model has the same hyperparameters as Medium (M) model.}
\label{tab:hyperparameters}
\end{table*}

\section{Results}
\label{sec:results}

We evaluate our model for image classification on ImageNet-1K (IN1K)~\citep{deng2009imagenet} and for video recognition on Kinetics-400 (K400)~\citep{kay2017k400} and Something-Something-v2 (SSv2)~\citep{goyal2017ssv2}. For evaluating the robustness to various common corruptions, we use the ImageNet-C (IN-C)~\citep{hendrycks2019corrImageNet} benchmark. Note that ImageNet-C is only used for testing, but not for training.


\subsection{Evaluation on ImageNet-1K}
\label{sec:results:image}
We use the IN1K~\citep{deng2009imagenet} dataset for pre-training our models. IN1K contains 1.28M training and 50k validation images for 1000 categories. The models pre-trained on IN1K are used as an initializing point for fine-tuning on the other datasets.

\textbf{Evaluation Setup:} We train our models for 300 epochs, using the AdamW optimizer~\citep{loshchilov2018adamw} with a learning rate of 5e-4, weight decay of 0.1, a batch size of 128 per GPU and input image resolution of 224 and a patch size of 16. We set the initial linear warm-up epochs as 5. We set the ratio of Transformer blocks to Mamba blocks to 1:7 for our baseline models. We use 4 nodes with 4 A100 GPUs (40GB) each for training. We do not use any automatic mixed precision. For a fair comparison, we also train our models with and without distillation to gauge the effect of distillation on the overall training scheme and architecture. The complete set of hyperparameters is provided in \cref{tab:hyperparameters}. 

\begin{table}[t!]
\centering
\resizebox{\textwidth}{!}{
    \begin{tabular}{l|l|c|c|c|c|c}
    \toprule
        \textbf{Type}& \textbf{Model} & \textbf{iso.} & \begin{tabular}[c]{@{}c@{}}\textbf{\makecell[c]{Image\\Size}}\end{tabular} & \textbf{\makecell[c]{\#Params\\(M)}} & \textbf{\makecell[c]{FLOPs\\(G)}} & \begin{tabular}[c]{@{}c@{}}\textbf{\makecell[c]{IN1K\\Top-1\%}}\end{tabular} \\
    \Xhline{0.5pt}
            \multirow{3}{*}{\textit{\textbf{CNN}}} & ConvNeXt-T~\citep{liu2022convnext} & \xmark & 224$^2$ & 29 & 4.5 & 82.1\Tstrut\\
            & ConvNeXt-S~\citep{liu2022convnext} & \xmark & 224$^2$ & 50 & 8.7 & 83.1 \\
            & ConvNeXt-B~\citep{liu2022convnext} & \xmark & 224$^2$ & 89 & 15.4 & 83.8 \\
    \Xhline{0.5pt}    
            \multirow{3}{*}{\textbf{\textit{\makecell[l]{CNN+\\SSM.}}}} & VMamba-T~\citep{yue2024vmamba} & \xmark & 224$^2$ & 31 & 4.9 & 82.2\Tstrut\\
            & VMamba-S~\citep{yue2024vmamba} & \xmark & 224$^2$ & 50 & 8.7 & 83.5 \\
            & VMamba-B~\citep{yue2024vmamba} & \xmark & 224$^2$ & 89 & 15.4 & 83.7 \\
    \Xhline{0.5pt} 
            \multirow{6}{*}{\textbf{\textit{\makecell[l]{Trans.}}}}& Swin-T~\citep{liu2021Swin} & \xmark & 224$^2$ & 28 & 4.6 & 81.3\Tstrut\\
            & Swin-S~\citep{liu2021Swin} & \xmark & 224$^2$ & 50 & 8.7 & 83.0 \\
            & Swin-B~\citep{liu2021Swin} & \xmark & 224$^2$ & 88 & 15.4 & 83.5 \\
            & DeiT-T~\citep{touvron2021deit} & \cmark & 224$^2$ & 6  & 1.3 & 72.2 \\
            & DeiT-S~\citep{touvron2021deit} & \cmark & 224$^2$ & 22 & 4.6 & 79.8 \\
            & DeiT-B~\citep{touvron2021deit} & \cmark & 224$^2$ & 87 & 17.6 & 81.8 \\
    \Xhline{0.5pt}
            \multirow{13}{*}{\textbf{\textit{\makecell[l]{SSM}}}} & ViM-T~\citep{lianghui2024vim} & \cmark & 224$^2$ & 7  & 1.1 & 76.1  \\
            & ViM-S~\citep{lianghui2024vim} & \cmark & 224$^2$ & 26  & 4.3 & 80.5 \\
            
            & VideoMamba-T~\citep{yue2024vmamba} & \cmark & 224$^2$ & 7 & 1.1 & 76.9 \\
            & VideoMamba-S~\citep{yue2024vmamba} & \cmark & 224$^2$ & 26 & 4.3 & 81.2 \\
            & VideoMamba-M~\citep{yue2024vmamba} & \cmark & 224$^2$ & 74 & 12.7 & 81.4 \\
        & VideoMamba-M$^\dagger$~\citep{yue2024vmamba} & \cmark & 224$^2$ & 74 & 12.7 & 82.8 \\
        & VideoMamba-B$^\dagger$~\citep{yue2024vmamba} & \cmark & 224$^2$ & 98 & 16.9 & 82.7 \\
      \rowcolor{\mycolor}
        \cellcolor{white}  & StableMamba-T & \cmark & 224$^2$ & 7 & 1.2 & 77.4 \\
        \rowcolor{\mycolor}
        \cellcolor{white}    & StableMamba-S & \cmark & 224$^2$ & 27 & 4.4 & 81.5 \\
        \rowcolor{\mycolor}
        \cellcolor{white}    & StableMamba-M & \cmark & 224$^2$ & 76 & 12.9 & 83.1 \\
        \rowcolor{\mycolor}
        \cellcolor{white}    & StableMamba-M$^\dagger$ & \cmark & 224$^2$ & 76 & 12.9 & 83.5 \\
        \rowcolor{\mycolor}
        \cellcolor{white}    & StableMamba-B & \cmark & 224$^2$ & 101 & 17.1 & 83.9 \\
        \rowcolor{\mycolor}
        \cellcolor{white}    & StableMamba-B$^\dagger$ & \cmark & 224$^2$ & 101 & 17.1 & 84.1 \\
    \bottomrule
    \end{tabular}}
    \normalsize
\caption{\label{tab:imagenet}\textbf{Performance comparison on ImageNet-1K:} We report the performance of our proposed models with state-of-the-art Mamba-based models and popular convolution-based and Transformer-based models on the ImageNet-1K~\citep{deng2009imagenet} validation set. Our proposed models outperform the Mamba-based models. 
$^\dagger$ represents the results using distillation. `iso.` means isotropic.}
\end{table}

\textbf{Results:} We present results for evaluating StableMamba on the IN1K dataset with other comparable methods in \cref{tab:imagenet}. We train our method with and without distillation to show the impact of distillation on the accuracy. We first compare the results without distillation. StableMamba outperforms the current state-of-the-art isotropic visual SSM models (ViM and VideoMamba) on IN1K for all model sizes. Compared to VideoMamba, the improvement ($+1.7$) of StableMamba is largest for the model M, which is largest model of VideoMamba that can be trained without distillation. Note that an improvement of $+1.7$ on IN1K is substantial. The improvements compared to VideoMamba are visualized by the solid lines in \cref{fig:only_intro}, which show the lack of scalability of VideoMamba. If we compare VideoMamba and StableMamba with distillation, we observe that distillation improves the accuracy for both architectures, but StableMamba still outperforms VideoMamba. The accuracy of StableMamba-B$^\dagger$ is $+1.4$ higher than of VideoMamba-B$^\dagger$. It is interesting to note that StableMamba-B without distillation even outperforms VideoMamba-B$^\dagger$ with distillation by $+1.2$. Most important, however, is that StableMamba can be scaled up and does not need any distillation as shown in \cref{fig:only_intro}.   


\subsection{Evaluation on Video Recognition}
\label{sec:results:video}

\begin{table}[tb]
\centering
\resizebox{\columnwidth}{!}{%
    \begin{tabular}{l|l|c|r|c|r|c}
    \toprule
        \multirow{1}*{\textbf{Arch.}} & \multirow{1}*{\textbf{Model}} &  \textbf{P.T.} & \textbf{\makecell[c]{Input\\Size}} & \textbf{\makecell[c]{\#Params\\(M)}} & \textbf{\makecell[c]{FLOPs\\(G)}} & \multicolumn{1}{c}{\textbf{\makecell[c]{K400\\Top-1\%}}} \\
        \Xhline{0.8pt}
         \multirow{4}{*}{\textit{\textbf{CNN}}} & SlowFast$_{R101+NL}$ & \multirow{2}{*}{-} & \multirow{2}{*}{80$\times$224$^2$} & \multirow{2}{*}{60} & \multirow{2}{*}{234$\times$3$\times$10} &\multirow{2}{*}{79.8}\Tstrut\\
         ~ & \citep{feichtenhofer2019slowfast} & & & & &\\
         ~ & X3D-M~\citep{feichtenhofer2020x3d}  & - & 16$\times$224$^2$ & 4 & 6$\times$3$\times$10 &76.0 \\
         ~ & X3D-XL~\citep{feichtenhofer2020x3d} & - & 16$\times$312$^2$ & 20 & 194$\times$3$\times$10 &80.4 \\
         \hline
         \multirow{5}{*}{\textbf{\textit{\makecell[l]{CNN+\\Trans.}}}} & MViTv1-B~\citep{fan2021multiscale} &  - & 32$\times$224$^2$ & 37 & 70$\times$1$\times$5 & 80.2\Tstrut\\
         ~ & MViTv2-S~\citep{li2022improved} &  - & 16$\times$224$^2$ & 35 & 64$\times$1$\times$5 & 81.0 \\
         ~ & UniFormer-S~\citep{li2022uniformer} &  IN1K & 16$\times$224$^2$ & 21 & 42$\times$1$\times$4 & 80.8 \\
         ~ & UniFormer-B~\citep{li2022uniformer} &  IN1K & 16$\times$224$^2$ & 50 & 97$\times$1$\times$4 & 82.0 \\
         ~ & UniFormer-B~\citep{li2022uniformer} &  IN1K & 32$\times$224$^2$ & 50 & 259$\times$3$\times$4 & 83.0\\
         \Xhline{0.8pt}
         \multirow{8}{*}{\textit{\textbf{Trans.}}} & Swin-T~\citep{liu2021video-swin} &  IN1K & 32$\times$224$^2$ & 28 & 88$\times$3$\times$4     & 78.8\Tstrut \\
         ~ & Swin-B~\citep{liu2021video-swin} & IN1K & 32$\times$224$^2$ & 88 & 88$\times$3$\times$4 & 80.6  \\
         ~ & Swin-B~\citep{liu2021video-swin} & IN21K & 32$\times$224$^2$ & 88 & 282$\times$3$\times$4 & 82.7 \\
         ~ & STAM (Sharir et al.~2021) &  IN21K & 64$\times$224$^2$ & 121 & 1040$\times$1$\times$1 & 79.2 \\
         ~ & TimeSformer-L & \multirow{2}{*}{IN21K} & \multirow{2}{*}{96$\times$224$^2$} & \multirow{2}{*}{121} & \multirow{2}{*}{2380$\times$3$\times$1} & \multirow{2}{*}{80.7} \\
         ~ & (Bertasius et al.~2021) & & &  &  & \\
         ~ & ViViT-L~\citep{arnab2021vivit} & IN21K & 16$\times$224$^2$ & 311 & 3992$\times$3$\times$4 & 81.3  \\
         ~ & Mformer-HR~\citep{patrick2021mformer} & IN21K & 16$\times$336$^2$ & 311 & 959$\times$3$\times$10 & 81.1  \\

         \hline
             \multirow{7}{*}{\textit{\textbf{SSM}}} & {VideoMamba-T~\citep{li2024videomamba} } &  {IN1K} &{16$\times$224$^2$} &{7} &{17$\times$3$\times$4} &78.1\Tstrut\\
             ~ &{VideoMamba-S~\citep{li2024videomamba} } &  {IN1K} &{16$\times$224$^2$} &{26} &{68$\times$3$\times$4} &80.8 \\
             ~ &{VideoMamba-M$^\dagger$~\citep{li2024videomamba} }& {IN1K} &{16$\times$224$^2$} &{74} &{202$\times$3$\times$4} &81.9 \\
         ~ & \cellcolor{\mycolor}{StableMamba-T} & \cellcolor{\mycolor}{IN1K} & \cellcolor{\mycolor}{16$\times$224$^2$} & \cellcolor{\mycolor}{7} & \cellcolor{\mycolor}{19$\times$3$\times$4} & \cellcolor{\mycolor}78.6 \\
         ~ & \cellcolor{\mycolor}{StableMamba-S} & \cellcolor{\mycolor}{IN1K} & \cellcolor{\mycolor}{16$\times$224$^2$} & \cellcolor{\mycolor}{27} & \cellcolor{\mycolor}{70$\times$3$\times$4} & \cellcolor{\mycolor}81.2\\
         ~ & \cellcolor{\mycolor}{StableMamba-M} & \cellcolor{\mycolor}{IN1K} & \cellcolor{\mycolor}{16$\times$224$^2$} & \cellcolor{\mycolor}{76} & \cellcolor{\mycolor}{206$\times$3$\times$4} & \cellcolor{\mycolor}82.2\\
      
      ~ & \cellcolor{\mycolor}{StableMamba-M$^\dagger$} & \cellcolor{\mycolor}{IN1K} & \cellcolor{\mycolor}{16$\times$224$^2$} & \cellcolor{\mycolor}{76} & \cellcolor{\mycolor}{206$\times$3$\times$4} & \cellcolor{\mycolor}82.5\\
    \Xhline{1.0pt}	
    \end{tabular}}
\caption{\label{tab:k400}Comparison with state-of-the-art methods on Kinetics-400~\citep{kay2017k400}. $^\dagger$ represents initialization with ImageNet-1K pretraining using distillation.}
\end{table}

\begin{table}[t!]
\centering
\resizebox{\columnwidth}{!}{
    \begin{tabular}{l|l|c|c|r|c}
    \toprule
        \multirow{1}*{\textbf{Arch.}} & \multirow{1}*{\textbf{Model}} & \textbf{P.T.} & \textbf{\makecell[c]{\#Params\\(M)}} & \textbf{\makecell[c]{FLOPs\\(G)}} & \multicolumn{1}{c}{\textbf{\makecell[c]{SSv2\\Top-1\%}}} \\
    \Xhline{0.8pt}
     \multirow{4}{*}{\textit{\textbf{CNN}}} & SlowFast$_{R101}$ & \multirow{2}{*}{K400} & \multirow{2}{*}{53} & \multirow{2}{*}{106$\times$3$\times$1} & \multirow{2}{*}{63.1}\Tstrut\\
     ~ & \citep{feichtenhofer2019slowfast} & &  & & \\
     ~ & CT-Net$_{R50}$~\citep{li2020ct-net} & IN1K & 21 & 75$\times$1$\times$1 & 64.5 \\
     ~ & TDN$_{R50}$~\citep{wang2021tdn}& IN1K & 26 & 75$\times$1$\times$1 &65.3  \\
     \Xhline{0.5pt}
     \multirow{6}{*}{\textbf{\textit{\makecell[l]{CNN+\\Trans.}}}} & MViTv1-B~\citep{fan2021multiscale}& K400 & 37 & 71$\times$3$\times$1 &64.7\Tstrut\\
     ~ & MViTv1-B~\citep{fan2021multiscale}& K400 & 37 & 170$\times$3$\times$1 &67.1  \\
     ~ & MViTv2-S~\citep{li2022improved}& K400 & 35 & 65$\times$3$\times$1 &68.2  \\
     ~ & MViTv2-B~\citep{li2022improved}& K400 & 51 & 225$\times$3$\times$1 &70.5\\
     ~ & UniFormer-S~\citep{li2022uniformer}& IN1K+K400 & 21 & 42$\times$3$\times$1 &67.7  \\
     ~ & UniFormer-B~\citep{li2022uniformer}& IN1K+K400 & 50 & 97$\times$3$\times$1 &70.4\\
     \Xhline{0.8pt}
     \multirow{5}{*}{\textit{\textbf{Trans.}}} & Swin-B~\citep{liu2021video-swin}& K400 & 89 & 88$\times$3$\times$1 & 69.6\Tstrut\\
     ~ & ViViT-L~\citep{arnab2021vivit}& IN21K+K400 & 311 & 3992$\times$3$\times$4 &65.4\\
     ~ & Mformer-HR~\citep{patrick2021mformer}& IN21K+K400 & 311 & 1185$\times$3$\times$1 &68.1\\
     ~ & TimeSformer-HR & \multirow{2}{*}{IN21K} & \multirow{2}{*}{121} & \multirow{2}{*}{1703$\times$3$\times$1} & \multirow{2}{*}{62.5}\\
     ~ & (Bertasius et al. 2021) & &  & & \\

    \Xhline{0.5pt}
    \multirow{7}{*}{\textit{\textbf{SSM}}} & VideoMamba-T~\citep{li2024videomamba} & IN1K   & 7 & 9$\times$3$\times$2 &65.1\Tstrut\\
    ~ & VideoMamba-S~\citep{li2024videomamba}  & IN1K   & 26 &  34$\times$3$\times$2 &66.6\\
    ~ & VideoMamba-M$^\dagger$~\citep{li2024videomamba}  & IN1K   & 74 &  101$\times$3$\times$4&67.3\\

      & \cellcolor{\mycolor}StableMamba-T    & \cellcolor{\mycolor}IN1K  & \cellcolor{\mycolor}7 &\cellcolor{\mycolor} 10$\times$3$\times$2 &\cellcolor{\mycolor}65.7\\

     & \cellcolor{\mycolor}StableMamba-S  & \cellcolor{\mycolor}IN1K  & \cellcolor{\mycolor}27 & \cellcolor{\mycolor} 35$\times$3$\times$2 &\cellcolor{\mycolor}67.3\\
    
     & \cellcolor{\mycolor}StableMamba-M   &  \cellcolor{\mycolor}IN1K  & \cellcolor{\mycolor}76 &\cellcolor{\mycolor} 103$\times$3$\times$4 &\cellcolor{\mycolor}67.8\\

    & \cellcolor{\mycolor}StableMamba-M$^\dagger$  &  \cellcolor{\mycolor}IN1K  & \cellcolor{\mycolor}76 & \cellcolor{\mycolor} 103$\times$3$\times$4 &\cellcolor{\mycolor}68.1\\
    \bottomrule

\end{tabular}}

\caption{\label{tab:ssv2}Comparison with state-of-the-art methods on the Something-Something-v2~\citep{goyal2017ssv2} dataset. $^\dagger$ represents initialization with ImageNet-1K pretraining using distillation. Network input sizes are the same as mentioned in K400.}
\end{table}

After pre-training on IN1K, we fine-tune the models on two large-scale datasets. The first dataset, K400~\citep{kay2017k400}, includes approximately 240,000 training videos and 19,000 validation videos, each about 10 seconds long, spanning 400 different human action classes. The second dataset, SSv2~\citep{goyal2017ssv2}, consists of around 220,000 videos: 168,000 for training, 24,000 for validation, and 27,000 for testing, covering 174 different classes.

\textbf{Evaluation Setup:} For fine-tuning, we use a batch size of 32 for tiny and a batch size of 16 for small variants due to the GPU memory limit. We set the number of linear warm-up epochs to 5, and the total number of epochs to 70 for K400 and 35 for SSv2 as in~\citep{li2024videomamba}. We use AdamW as an optimizer and a learning rate of 4e-4. The complete list of hyperparameters for reproducibility is provided in \cref{tab:hyperparameters}.    

\textbf{Results:} StableMamba demonstrates superior performance in downstream video recognition tasks compared to VideoMamba, which is the only Mamba architecture that can be applied to videos. On the K400 dataset in \cref{tab:k400}, StableMamba tiny and small outperform their VideoMamba counterparts without distillation. Distillation improves the accuracy for the middle models, but even with distillation StableMamba-M$^\dagger$ improves the accuracy of VideoMamba-M$^\dagger$ by $+0.6$, which is a substantial improvement on this dataset. The results on the SSv2 dataset shown in \cref{tab:ssv2} are similar, but the improvements are even larger. StableMamba-M$^\dagger$ improves the accuracy of VideoMamba-M$^\dagger$ by $+0.8$.


\subsection{Evaluation on ImageNet-C}
\label{sec:results:robustness}
IN-C~\citep{hendrycks2019corrImageNet} is a benchmark for evaluating the robustness of neural networks to images with common corruptions like JPEG compression. It includes 19 common types of image corruption at 5 different intensity levels. We test our network on this benchmark to assess the robustness introduced by attention layers. 

\textbf{Results:} We present results for Gaussian blurring and JPEG compression corruption for StableMamba-M in comparison with VideoMamba-M, ViT-B\textbackslash16 and ResNet-50 in \cref{fig:jpeg_compression}. We see that StableMamba-M (blue) outperforms VideoMamba-M (yellow) for all levels of corruption. The gap becomes larger as the intensity of corruption increases. StableMamba behaves similar or even slightly better than the pure attention-based architecture ViT-B\textbackslash16 and is more robust than ResNet-50, in particular for the highly relevant JPEG compression setting. 

We also report the results across all corruptions in the Table~\ref{tab:imagenet_c_mce_tab}. The Mean Corruption Error (mCE) table on the ImageNet-C dataset presented in Table ~\ref{tab:imagenet_c_mce_tab} showcases the robustness of various models to common image corruptions, with errors reported relative to AlexNet. Our proposed model, StableMamba-M, demonstrates superior performance with an mCE of 50.5\%, which is competitive with the DeiT-B model, which has an mCE of 50.4\%. Notably, StableMamba-M outperforms ViT-B/16 and VideoMamba-M, which have mCEs of 53.7\% and 51.6\%, respectively, highlighting its improved robustness. This comparison underscores StableMamba-M's effectiveness in enhancing model stability and corruption resistance, providing a significant advancement over existing models like VideoMamba.

\begin{table}[!ht]
    \centering
    \begin{tabular}{lcc}
    \toprule
        \multirow{1}*{\textbf{Model}} & \multirow{1}*{\textbf{Error on Clean}} & \multirow{1}*{\textbf{\makecell[b]{Mean Corruption Error\\(mCE)}}}\\ 
        \Tstrut\\
        \midrule
        AlexNet & 43.48\% & 100.0\% \\ 
        SqueezeNet1.1 & 41.82\% & 104.4\% \\ 
        VGG11 & 30.98\% & 93.5\% \\ 
        VGG19 & 27.62\% & 88.9\% \\ 
        VGG19BN & 25.78\% & 81.6\% \\ 
        DenseNet121 & 25.57\% & 73.4\% \\ 
        DenseNet169 & 24.40\% & 69.4\% \\ 
        DenseNet201 & 23.10\% & 68.4\% \\ 
        DenseNet161 & 22.86\% & 66.4\% \\ 
        CondenseNet4 & 26.25\% & 80.8\% \\ 
        CondenseNet8 & 28.93\% & 84.6\% \\ 
        ResNet18 & 30.24\% & 84.7\% \\ 
        ResNet34 & 26.69\% & 77.9\% \\ 
        ResNet50 & 23.87\% & 76.7\% \\ 
        ResNet101 & 22.63\% & 70.4\% \\ 
        ResNet152 & 21.69\% & 69.3\% \\ 
        ResNeXt50 & 22.89\% & 68.2\% \\ 
        ResNeXt101 & 21.81\% & 63.6\% \\ 
        ResNeXt101\_64 & 21.04\% & 62.2\% \\ 
        ViT-B/16 & 22.10\% & 53.7\% \\ 
        DeiT-B & 18.20\% & 50.4\% \\
        VideoMamba-M & 18.60\% & 51.6\% \\
        \cellcolor{\mycolor}StableMamba-M & \cellcolor{\mycolor}16.90\% & \cellcolor{\mycolor}50.5\% \\ 
        \bottomrule
    \end{tabular}
    
    \caption{\label{tab:imagenet_c_mce_tab} Mean Corruption Error (mCE) on ImageNet-C~\citep{hendrycks2019corrImageNet} dataset across all 19 corruptions. mCE is reported relative to AlexNet~\citep{alex2012alexnet} errors on ImageNet-C.}
\end{table}


\subsection{Ablation Studies}
\label{sec:results:ablation}

\begin{figure}[H]
    \centering
    \includegraphics[width=\columnwidth]{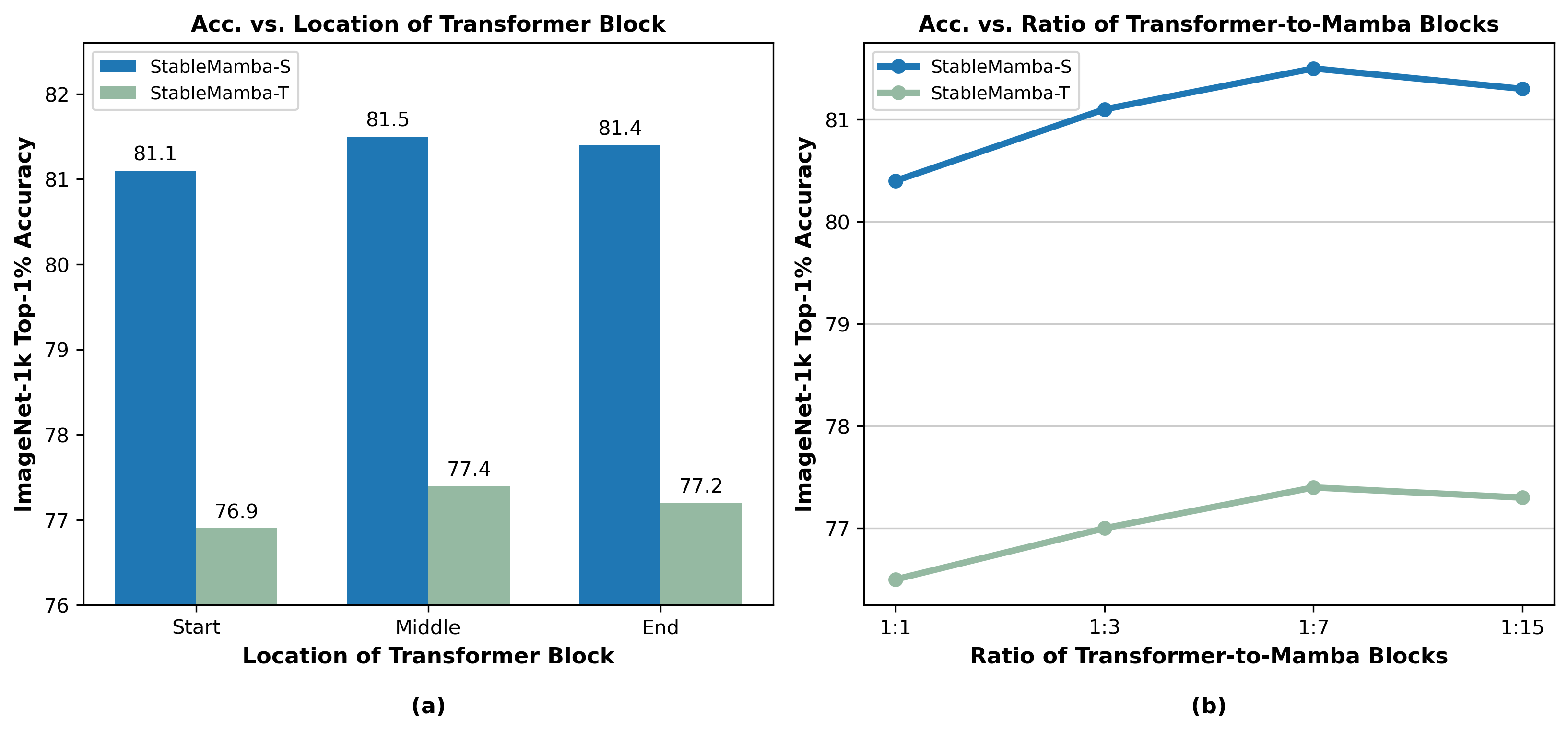}
    \caption{\label{fig:ablation}\textbf{(a)} Impact of the position of the Transformer block within StableMamba. \textbf{(b)} Impact of the ratio of Transformer blocks to Mamba blocks.}
    
\end{figure}

\textbf{Position of Transformer block:} In \cref{fig:overall_architecture}(a), the Transformer block is placed in the middle of the StableMamba blocks. This position results from our analysis of the impact on the location of the Transformer block. We conducted three experiments each for StableMamba-T and StableMamba-S, totaling six experiments, to determine the optimal position for the Transformer block. We tested placing the Transformer block at the start, middle, and end of the StableMamba blocks and evaluated their performance on the IN1K dataset. As shown in \cref{fig:ablation}(a), the performance of StableMamba is not highly sensitive to the Transformer's position in both tiny and small models. However, there is a slight performance improvement when the Transformer block is in the middle.

Therefore, we use the middle position as the default for our StableMamba architecture.

\textbf{Number of Transformer Blocks:} Similar to the position of Transformer blocks within each StableMamba block, the ratio of Transformer blocks to Mamba blocks is another design parameter for the StableMamba block. We interleave a Transformer block for every $k$ Mamba block; for example, we interleave one Transformer block for every seven Mamba blocks. To evaluate the impact of the ratio, we conducted experiments varying the number of Mamba blocks per Transformer block. As shown in \cref{fig:ablation}(b), the performance on the IN1K dataset improves as the number of Mamba blocks per Transformer block increases, reaching optimal accuracy at a ratio of 1:7. Beyond this ratio, the performance decreases. Therefore, we set the design parameter to one Transformer block for every seven Mamba blocks in the StableMamba architecture.

\textbf{Dependence on context length:}
Apart from the network architecture itself, it is interesting to investigate the network with context lengths of different sizes. To probe the suitability of our approach for a long context, we perform additional experiments. First, we train StableMamba-T with a longer context for video classification, using 32 frames instead of the usual 16 frames. Second, we train StableMamba with a larger resolution (448 instead of 224) to see its effect on image classification as well. The results in \cref{tab:long_context_ablation} show that StableMamba and VideoMamba benefit from the increased context length, which is a general strength of Mamba-based architectures. In all cases, StableMamba outperforms VideoMamba.    

\begin{table}[h]
\centering
\begin{tabular}{l|c|c|c|c}

\textbf{\makecell[c]{Model}} & \textbf{\makecell[c]{Context\\Length}} & \textbf{\makecell[c]{Training\\Dataset}}  & \textbf{\makecell[c]{FLOPs\\(G)}}  & \textbf{Accuracy} \\ \Xhline{0.8pt}
VideoMamba-T   & $224^2$ & IN1K  & 1.1 & 76.9\%\Tstrut          \\
StableMamba-T  & $224^2$ & IN1K  & 1.2 & \textbf{77.4\%} \\
VideoMamba-T   & $448^2$ & IN1K  & 4.3 & 79.3\%          \\
StableMamba-T  & $448^2$ & IN1K  & 4.5 & \textbf{79.9\%} \\ \Xhline{0.8pt}
VideoMamba-T   & $16\times224^2$ & K400 & $17\times3\times4$ & 78.1\%\Tstrut          \\
StableMamba-T   & $16\times224^2$ & K400 & $19\times3\times4$ & \textbf{78.6\%} \\
VideoMamba-T    & $32\times224^2$ & K400 & $34\times3\times4$ & 78.8\%          \\
StableMamba-T   & $32\times224^2$ & K400 & $37\times3\times4$ & \textbf{79.3\%} \\
\end{tabular}
\caption{Impact of image resolution (top) and number of input frames (bottom)  for StableMamba and VideoMamba.}
\label{tab:long_context_ablation}
\end{table}

\begin{figure}[H]
    \centering
    \includegraphics[width=\columnwidth]{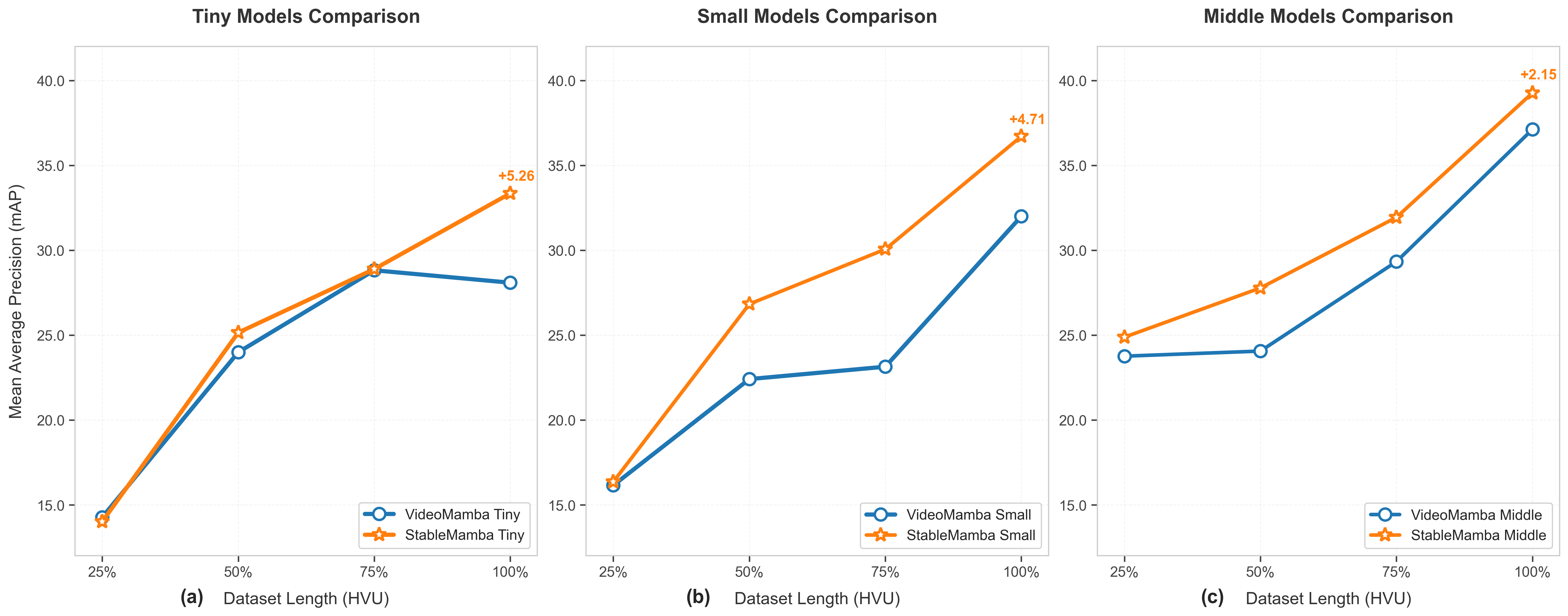}
    \caption{\label{fig:hvu}\textbf{(a)} Dataset scaling experiment using 25\%, 50\%, 75\%, and 100\% of the training dataset while performing the validation on the full validation set.}
    
\end{figure}

\textbf{Dependence on dataset length:}
Along with the context length, it is also interesting to ablate data efficiency of the network. For this purpose we conducted scaling experiments using 25\%, 50\%, 75\%, and 100\% of the training dataset while performing the validation on the full validation set. Results (in \cref{fig:hvu}) show our network consistently outperforms VideoMamba model across all data regimes. While conventional approaches exhibit performance saturation as data volume increases, our architecture maintains higher accuracy at each threshold and continues to improve with additional data. The performance gap is already evident at the 25\% level for small and middle model and progressively widens with dataset scaling, confirming that our modifications enable better feature extraction from limited samples without compromising the ability to leverage larger datasets.

\section{Conclusion}
\label{sec:conclusion}
We have investigated and addressed the scalability challenge in large visual state-space models by proposing a straightforward interleaved design that scales effectively to a substantial number of parameters, consistently outperforming smaller models. Our ablation studies provide insights regarding optimal positioning, the number of attention layers in the architecture, and its robustness to common corruptions in the input like JPEG compression. Extensive experiments show that our method enables the scaling of Mamba-based models to over 100M parameters, significantly enhancing performance while also improving overall robustness. Evaluations on the K400 and SSv2 datasets for video recognition validate that our approach achieves state-of-the-art results.

\section*{Acknowledgements}
The work has been supported by the Federal Ministry of Education and Research (BMBF) under grant no. 01IS22094E WEST-AI and the ERC Consolidator Grant FORHUE (101044724). The authors gratefully acknowledge the Gauss Centre for Supercomputing e.V.
(www.gauss-centre.eu) for funding this project by providing computing time through the John von Neumann
Institute for Computing (NIC) on the GCS Supercomputer JUWELS at Jülich Supercomputing Centre
(JSC). The authors also gratefully acknowledge EuroHPC Joint Undertaking for awarding us access to Leonardo at CINECA, Italy, through EuroHPC Regular Access Call - proposal No.\ EHPC-REG-2024R01-076. The authors also gratefully acknowledge the granted access to the Marvin cluster hosted by the University of Bonn. 

\textbf{Funding:}
The work has been supported by the Federal Ministry of Education and Research (BMBF) under grant no. 01IS22094E WEST-AI and the ERC Consolidator Grant FORHUE (101044724).


\bibliography{sn-bibliography.bib}

\end{document}